# DialogueBERT: A Self-Supervised Learning based Dialogue Pre-training Encoder

Zhenyu Zhang
JD AI, Chengdu, China
zhangzhenyu47@jd.com

Tao Guo
Xiaoduo AI, Chengdu, China
guotao@xiaoduotech.com

Meng Chen
JD AI, Beijing, China
chenmeng20@jd.com

## ABSTRACT

With the rapid development of artificial intelligence, conversational bots have became prevalent in mainstream E-commerce platforms, which can provide convenient customer service timely. To satisfy the user, the conversational bots need to understand the user's intention, detect the user's emotion, and extract the key entities from the conversational utterances. However, understanding dialogues is regarded as a very challenging task. Different from common language understanding, utterances in dialogues appear alternately from different roles and are usually organized as hierarchical structures. To facilitate the understanding of dialogues, in this paper, we propose a novel contextual dialogue encoder (i.e. DialogueBERT) based on the popular pre-trained language model BERT. Five self-supervised learning pre-training tasks are devised for learning the particularity of dialouge utterances. Four different input embeddings are integrated to catch the relationship between utterances, including turn embedding, role embedding, token embedding and position embedding. DialogueBERT was pre-trained with 70 million dialogues in real scenario, and then fine-tuned in three different downstream dialogue understanding tasks. Experimental results show that DialogueBERT achieves exciting results with 88.63% accuracy for intent recognition, 94.25% accuracy for emotion recognition and 97.04% F1 score for named entity recognition, which outperforms several strong baselines by a large margin.

## 1 INTRODUCTION

Nowadays, conversational bots are widely applied in mainstream online shopping platforms such as Jingdong, Taobao, and Amazon. In the Chinese leading online shopping platform Jingdong, over 70% of the customers are served by the smart shopping assistant AlphaSales[1] for E-commerce customer service, covering user's various queries such as product questions, shipping policies, and promotion activities etc. Language understanding in dialogues is a challenging problem. To provide good user experience, the conversational bot needs to understand the intent of users in dialogues, recognize the entities to fill slots in conversations, and detect the user's emotion to generate empathetic responses. Unlike regular language understanding, conversational utterances appear alternately from different parties and are usually organized as hierarchical structures. The underlying difference of linguistic patterns between general text and dialogue utterances makes existing language understanding approaches less satisfying.

To understand the dialogue utterances accurately, lots of efforts have been made by devising various dialogue encoders and remarkable progresses have been achieved recently. One line of researches leverage the memory-networks [7, 14, 16, 19] to encode the dialogues for downstream tasks such as intent recognition, emotion recognition, etc. With the great success of pre-trained language models such as BERT [3, 4], RoBERTa [9], etc., some researchers transformed the dialogue modeling problem into a Machine Reading Comprehension (MRC) problem [1, 2, 12]. Recently, some dialogue-specific pre-training tasks are also proposed and obtain state-of-the-art (SOTA) results in the dialogue relevant downstream tasks. Qu et al. [10] incorporate history utterance embeddings into BERT model to find answer for user's question. The PT-CoDE [6] model leverages a hierarchical neural network to encode the conversation and pre-trains the model with conversation completion tasks. The TOD-BERT [18] pre-trains the BERT model from scratch by using both masked language modeling and response contrastive loss. DialoGPT [21] generates conversation responses by pre-training GPT2 [11] with huge amount dialogue corpus, which is mainly dedicated to response generation task instead of language understanding. Those pre-trained language models in open-domain and dialogue-specific domain all follow the self-supervised learning (SSL) [8] paradigm, where the input text is supposed to be recovered from partially observed input context.

Although the previous researches have proposed several SSL methods to pre-train the dialogue encoder, the dialogue-specific SSL is far from well-explored. There is still lack of enough researches for exploration of new SSL methods and comparison between different SSL pre-training tasks for dialogue encoding. To address this issue, in this paper, we propose a novel dialogue encoder (denoted as **DialogueBERT**) based on the prevalent pre-trained language model BERT [4]. To catch the particularity of dialogue utterances, we devise five self-supervised learning based pre-training tasks and train the DialogueBERT with large-scale conversation corpus. Inspired by [22], a convolutional pooler based on convolutional neural network (CNN) and 2D pooling is designed to extract the dialogue feature representations. To learn the relationship between different utterances in the dialogue, four different input embeddings are integrated, including token embedding, position embedding, role embedding and turn embedding. DialogueBERT is pre-trained with over 70 million of dialogues, which are collected from real conversations between users and customer service staffs in E-commerce scenario. To verify the effectiveness of our pre-trained dialogue encoder, extensive experiments were conducted by fune-tuning DialogueBERT on three downstream dialogue understanding tasks, including intent recognition, emotion recognition, and named entity recognition. Experimental results show that the proposed method outperforms several strong dialogue encoding approaches and achieves 88.63% accuracy score for intent recognition, 94.25% accuracy for emotion recognition and 97.04% F1 score for named entity recognition. Ablation study was also conducted to figure out the contribution of each SSL pre-training task, which can further inspire the future researches to build better dialogue encoder.

---

[1] AlphaSales is an intelligent conversational bot for answering customers' inquiries during shopping scenario, https://xiaozhi.jd.com/.

Specifically, our contributions include:

- We propose a novel pre-trained dialogue encoder based on BERT by devising five self-supervised learning pre-training tasks, including masked language modeling, masked utterance prediction, utterance replacement, turn swapping, and response selection. The model was pre-trained with more than 70 million of dialogues.
- We verify the effectiveness of DialogueBERT on three representative downstream dialogue tasks, including intent recognition, emotion recognition, and named entity recognition. Experimental results show the superiority and competitiveness of our proposed model.

## 2 APPROACH

In this Section, we will introduce the model input, model structure and self-supervised learning based model pre-training. Figure 1 demonstrates the overall architecture of our proposed model.

### 2.1 Model Input

In order to pre-train the model, we firstly collect a large-scale dialogue corpus from real conversations between customers and customer service staffs. After that, we desensitized and anonymized the private information based on very detailed rules. Finally, We collected over 70 million conversations with 8.59 utterances on average for each conversation. For each conversation utterance, we prepend one of the two special tags, i.e. "<Q>" (customer's query) and "<A>" (customer service staff's response). Then we concatenate the utterances to form the input text as "<Q>xxxx<A>xxx<Q>xxxx...". We trained a BPE sentencepiece tokenizer [13] with 5 million randomly sampled conversations. The vocabulary size is set to 50,000. Then we split the conversation input text into tokens. To catch the relationship between utterances, we combine the token, role and turn information for model input. As shown in Figure 1, tokens (denoted as "$T_1,T_2$", etc.) in each utterance are assigned with the corresponding turn information (indexed from "1" for the first utterance) and role information ("q" represents question and "a" denotes response). We restrict the maximum utterance length to 15 tokens and maximum conversation input text length to 128 tokens, where extra tokens are clipped.

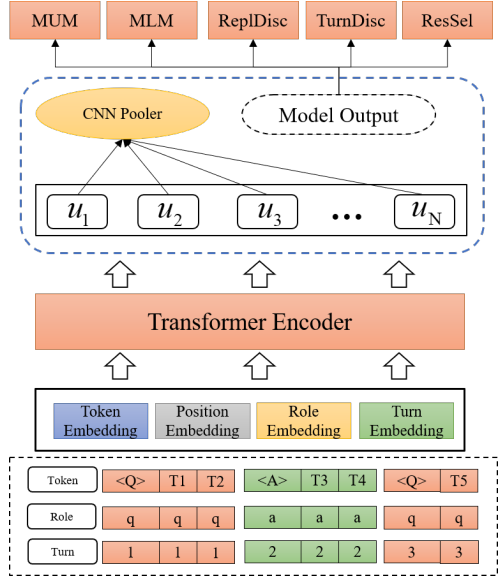

Figure 1: Overview architecture of DialogueBERT.

### 2.2 Model Structure

The proposed dialogue encoder is illustrated in Figure 1 which consists of four input embeddings, the Transformer Encoder, the Model Output and five SSL pre-training headers. The embedding layer consists of token embedding, position embedding, role embedding and turn embedding that can map the token, token position, role information and turn information of input sequence into dense vectors (i.e. $vec_{token}$, $vec_{pos}$, $vec_{role}$ and $vec_{turn}$ respectively). The embedding layer then sums up the four embeddings together and apply a layer normalization to obtain the final input embedding $vec_{emb}$.

$$vec_{emb} = vec_{token} + vec_{pos} + vec_{role} + vec_{turn} \qquad (1)$$

The Transformer Encoder is same as the BERT that contains 12 Transformer blocks [15]. It takes $vec_{emb}$ of each token as input and outputs $N$ contextual vectors (denoted as $u_1, u_2, ...u_N$ and $u_i \in R^d$, $1 \leq i \leq N$). Besides the contextual vectors, the model output also contains a dialogue representation feature vector. We use a CNN pooler to get the dialogue representation. There are six 1-D convolution kernels (filter sizes are 5, 7, 9, 11, 15, 20, where the output channel size is the same as BERT, i.e. each kernel $\in R^{K \times d}$ and $d$ is the dimension of 768) in the CNN. Inspired by [22], we apply a 2D-pooling on the CNN output to get the dialogue encoder pooling feature vector ($vec_{diag} \in R^d$). While the self-attention mechanism of BERT can capture the global semantics of input sequence, the convolution kernels applied along the input sequence dimension can capture the local ngram-level lexical and semantic information. The five SSL headers are used to pre-train the DialogueBERT with the huge conversation corpus, which will be described in detail in Section 2.3.

## 2.3 SSL-based Model Pre-Training

Considering the dialogue involves the token/word level feature, utterance level feature, and conversation order feature, etc., we design five different SSL pre-training tasks. As shown in Figure 1, the proposed SSL pre-training tasks are as follows:

- **MUM** is a masked utterance modeling SSL task, where we randomly mask one utterance in a conversation with several special tokens, i.e. "[MASK]" (replace the $m$ tokens in the utterance), then the model is required to generate the tokens of the masked utterance. The SSL method of the work in [6] selects the utterance from a candidate set (we denote this as **MSM**) while we directly predict the utterance in token level. We argue that because the candidate set is randomly sampled in [6], it is not a good choice for a contrastive SSL task due to poor contrastive samples [8]. We feed the masked token vectors $(u_{mask_1}, u_{mask_2}, ... u_{mask_m})$ to the masked language header to predict the original tokens by following the approach in BERT [4].
- **MLM** is a masked language modeling SSL task, where we follows the setting in [3]. This SSL task aims to learn the language modeling structure of the conversation text.
- **ReplDisc** randomly (with probability of 0.5) replaces one utterance in a conversation with a randomly selected utterance from another conversation in the same training batch, and then discriminates whether the new conversation is the replaced one or the original one. This SSL task is devised to learn the contextual feature of the conversation.
- **TurnDisc** randomly swaps two utterances in a conversation and then discriminates whether the new conversation is swapped, where the turn in new conversation is still indexed in ascending order (we do not swap the turn information of the two utterances). This SSL task is designed to learn the utterance order feature of conversation.
- **ResSel** is the same as the response constrastive loss (RCL) in [18], which can help learning a better representation for the [CLS] token, and capturing underlying dialogue sequential order, structure information, and response similarity. Differently, we only use the model to select the last utterance of a conversation from a set of candidate responses that are from the training batch.

For **ReplDisc** and **TurnDisc**, we use the logistic regression to predict the results with $vec_{diag}$ (Section 2.2) as input feature vector. The cross entropy is applied to compute loss for each SSL task. Then we sum up all the losses of the five SSL tasks in pre-training and apply LAMB optimizer [20] with maximum learning rate as 1e-4, warm-up steps as 10K and batch size as 256. We trained the DialogueBERT for 10 epochs with all the 70 million conversation corpus. For the downstream NLP tasks, we follow the settings of common language modeling task [17] implementation[2] to form them as sequence labelling task or sequence classification task, where the pooling vector $vec_{diag}$ (Section 2.2) is used for prediction.

[2]https://github.com/huggingface/transformers

| Datasets | Train | Dev | Test |
|---|---|---|---|
| IntR | 185.2K | 1K | 9.8K |
| EmoR | 208K | 1K | 11K |
| NER | 44.6K | 1K | 2.4K |

Table 1: The statistics of the experimental datasets

## 3 EXPERIMENTS

### 3.1 Datasets

To evaluate the DialogueBERT in downstream conversation understanding tasks, we experiment on three representative tasks, including intent recognition, emotion recognition, and named entity recognition. All experiments were conducted on our in-house datasets[3]. Table 1 shows the statistics of the datasets, where the term "IntR", "NER" and "EmoR" refer for intent recognition, named entity recognition and emotion recognition respectively. We split the data into 95% for training to fine-tune the model and 5% for testing randomly. For the training set, we hold out 1000 samples as validation set to select the best model in fine-tuning. There are 102 intents, 5 emotions (angry, satisfied, sad, curious, neural) and 28 kinds of named entities in the corresponding datasets.

For intent recognition and emotion recognition, we feed the encoder pooling feature vector ($vec_{diag} \in R^d$, Section 2.2) from CNN pooler (Figure 1) into a following classification head $W \in R^{d \times C}$ ($C$ is the number of intents or emotions, i.e. C is 102 and 5 for IntR and EmoR respectively) for prediction. We get the predictions via $W vec_{diag} \in R^C$ and calculate the classification loss via Equation 2 as follows:

$$loss^{class} = -log(softmax(W vec_{diag})) \quad (2)$$

For the named entity recognition (NER), we use the BIO tagging schema to create the annotated corpus, where $B$ represents the beginning of a named entity, $I$ represents the inside tokens of a named entity and $O$ represents the non-entity tokens. We use the $N$ output vectors of input tokens, denoted as $U \in R^{N \times d}$ ($U = \{u_1, u_2, ..., u_N\}$), and then apply a standard token classification head ($W^{ner} \in R^{d \times P}$) to get the token-level classification output $W^{ner}U \in R^{N \times P}$, where $P$ is the number of tagging labels (i.e. 28 B-ne tags, 28 I-ne tags and 1 O tag, totally 28 × 2 + 1). We compute the NER loss via Equation 3 during fine-tuning.

$$loss^{ner} = -\frac{1}{N} \sum_{i=1}^{i=N} log(softmax(W^{ner}u_i)), 1 \leq i \leq N \quad (3)$$

### 3.2 Experiment Setup

We evaluate the effectiveness of DialogueBERT in three tasks: intent recognition (IntR), named entity recognition (NER) and emotion recognition (EmoR). For IntR and EmoR, we use the *classification*

[3]We are working on data desensitization and the datasets will soon be released to the public to facilitate future research.

| Model | IntR | NER | EmoR |
|---|---|---|---|
| DMN [7] | 83.15 | - | 90.16 |
| KVNet [19] | 74.22 | - | 85.47 |
| PT-CoDE [6] | 86.36 | - | 92.95 |
| TOD-BERT [18] | 87.91 | 96.85 | 92.83 |
| $DialogueBERT_{CLS}$ | 88.32 | 97.00 | 94.16 |
| **DialogueBERT** | **88.63** | **97.04** | **94.25** |

Table 2: Performance comparison of different models on three downstream dialogue understanding tasks.

| SSL task | IntR | NER | EmoR |
|---|---|---|---|
| MUM | 87.92 | 96.95 | 93.56 |
| MSM | 87.63 | 96.59 | 92.65 |
| ResSel | 87.97 | 96.85 | 92.88 |
| ReplDisc | 87.24 | 96.11 | 92.15 |
| TurnDisc | 86.25 | 96.03 | 92.07 |
| **DialogueBERT** | **88.63** | **97.04** | **94.25** |

Table 3: Ablation Study of each SSL pre-training task. MLM task is combined to each pre-training task by default for stable training.

*accuracy* as evaluation metric. For NER, the *Macro-F1 score* is reported as evaluation metric. We compared our method with the following four strong baselines:

- **DMN** [7]: the dynamic memory network formulates dialogue understanding as QA tasks, which updates the memory iteratively with a dynamic gate via attention mechanism.
- **KVNet** [19]: it applied key-value memory neural network as a semantic parsing module to approach the open-domain KB-QA task, which selects relevant history utterances to generate a feature vector.
- **PT-CoDE** [6] is a hierarchical dialogue encoder and is pretrained with a conversation completion task (CoCo).
- **TOD-BERT** [18] learns the dialogue structure and utterance order via both masked language modeling loss and response contrastive loss.

As to the implementation details, we leveraged the works of Huggingface[2] [17] and texar[4] [5] to implement those models. We set the model dimension as 768 for DialogueBERT and all the baselines.

### 3.3 Experimental Results

As shown in Table 2, DialogueBERT outperforms the baselines in all the three downstream dialogue understanding tasks. Our method gets **88.63%** accuracy score for intent recognition, **97.04%** Macro-F1 score for named entity recognition and **94.25%** accuracy for emotion recognition, which outperforms the TOD-BERT by 0.72%, 0.19% and 1.42% respectively (statistically significant difference with $p < 0.05$). It's also observed that all the SSL based methods (TOD-BERT, PT-CoDE, DialogueBERT) outperform the memory networks based methods significantly, which demonstrate the superiority of SSL methods in dialogue encoding.

The $DialogueBERT_{CLS}$ in Table 2 denotes that we prepend a special classification token "[CLS]" token to the conversation input text and use the transformer encoder output $u_{CLS}$ as pooling vector rather than applying our CNN pooler (Section 2.2). The "CLS" pooling strategy is previously used by the open-domain BERT [4] and dialogue-specific encoder TOD-BERT [18] (TOD-BERT outperforms the original open-domain BERT in dialogue-specific tasks). Compared with the DialogueBERT, the performance of $DialogueBERT_{CLS}$ drops 0.31%, 0.04% and 0.09% for intent recognition, named entity recognition and emotion recognition respectively while it still outperforms the other baselines. It indicates that, on the one hand, the performance gain of DialogueBERT is mainly from the five SSL tasks. Meanwhile, it also demonstrates the contribution of our CNN pooler.

### 3.4 Ablation Study

Table 3 shows the experimental results for each SSL pre-training task. We found that the model is hard to converge without MLM, which results in a poor pre-trained dialogue encoder that could barely fit any downstream task. Therefore, we report the model performance for each SSL task together with MLM. Because the research in [18] has proved that MLM and ResSel based method outperforms the pure MLM based pre-training method, here we skip repeating the similar experiments for space limitation.

We found that the MUM is the most effective SSL pre-training task, which can learn the dialogue structure to facilitate both the token level and dialogue level pre-training. The ResSel achieves similar performance with MUM but better score than MSM, which demonstrates that the last utterance in the conversation is very important for building the dialogue encoder. Please note that the research in [6] applies the MSM (conversation completion; refer to Section 2.2) for pre-training, while our experiments show that the MUM SSL method is better for dialogue encoder. The MUM outperforms the MSM by 0.29%, 0.36% and 0.91% in the three tasks (i.e. IntR, NER, EmoR) respectively. We argue that because there are usually within 15 tokens for each utterance, it is easier for the model to generate the utterance in token level, where the model can avoid suffering from poor constrastive candidate samples in [6]. Overall speaking, each of the proposed SSL task contributes to the final gain of DialogueBERT to some degree. Considering the different contributions for each task, we will try to combine the five losses with different weights in the future.

## 4 CONCLUSION AND FUTURE WORK

In this paper, we propose a novel dialogue pre-training encoder, i.e. DialogueBERT. Five different self-supervised learning based pre-training tasks are devised to catch the particularity of conversation utterances and enhance the dialogue representations. Then we verify the effectiveness of DialogueBERT on three representative downstream dialogue understanding tasks, including intent recognition, named entity recognition and emotion recognition. We also analyze the contribution of each SSL task independently. In the future, we plan to design query rewriting task in the pre-training stage to enhance the context modeling.

---
[4]https://github.com/asyml/texar